\documentclass[a4paper,12pt]{IEEEtran}
\usepackage{cite}
\usepackage{graphicx,epsfig,amssymb,multirow}
\usepackage[cmex10]{amsmath}
\usepackage{algorithmic,algorithm}
\usepackage[tight,normalsize,sf,SF]{subfigure}

\begin{document}
%
\title{A Fusion Approach for \\ Efficient Human Skin Detection}

\author{Wei~Ren~Tan,
Chee~Seng~Chan,
Pratheepan~Yogarajah
and~Joan~Condell
\thanks{~~~~W.R. Tan and C.S. Chan are with the Centre of Image and Signal Processing, Faculty of Computer Science and Information Technology, University of Malaya, 50603 Kuala Lumpur, Malaysia (e-mail: willtwr@siswa.um.edu.my; cs.chan@um.edu.my).}
\thanks{~~~~P. Yogarajah and J. Condell are with the School of Computing and Intelligent Systems, University of Ulster (Magee), Northern Ireland, BT48 7JL, U.K. (e-mail: p.yogarajah@ulster.ac.uk; j.condell@ulster.ac.uk).}}
\markboth{Accepted in IEEE Transactions on Industrial Informatics, Vol.8(1), pp. 138-147}%
{Shell \MakeLowercase{\textit{et al.}}: Bare Demo of IEEEtran.cls for Computer Society Journals}
%
\IEEEtitleabstractindextext{%
\begin{abstract}
A reliable human skin detection method that is adaptable to different human skin colours and illumination conditions is essential for better human skin segmentation. Even though different human skin colour detection solutions have been successfully applied, they are prone to false skin detection and are not able to cope with the variety of human skin colours across different ethnic. Moreover, existing methods require high computational cost. In this paper, we propose a novel human skin detection approach that combines a smoothed 2D histogram and Gaussian model, for automatic human skin detection in colour image(s). In our approach an eye detector is used to refine the skin model for a specific person. The proposed approach reduces computational costs as no training is required; and it improves the accuracy of skin detection despite wide variation in ethnicity and illumination. To the best of our knowledge, this is the first method to employ fusion strategy for this purpose. Qualitative and quantitative results on three standard public datasets and a comparison with state-of-the-art methods have shown the effectiveness and robustness of the proposed approach.
\end{abstract}
\begin{IEEEkeywords}
Skin Detection, Dynamic Threshold, Colour Space, Fusion Strategy
\end{IEEEkeywords}}

\maketitle

\IEEEdisplaynontitleabstractindextext

%
\IEEEpeerreviewmaketitle

\section{Introduction}
With the progress of information society today, images have become more and more important. Among them, skin detection plays an important role in a wide range of image processing applications from face tracking, gesture analysis, content-based image retrieval systems to various human-computer interaction domains \cite{Vadakkepat,Chan2,Kubota,Linda,Myself,Pratl}. In these applications, the search space for objects of interests, such as hands, can be reduced through the detection of skin regions. 
One of the simplest and commonly used human skin detection methods is to define a fixed decision boundary for different colour space components \cite{Sobottka, HyeonBae,Wang}. Single or multiple ranges of threshold values for each colour space components are defined and the image pixel values that fall within these pre-defined range(s) are selected as skin pixels. In this approach, for any given colour space, skin colour occupies a part of such a space, which might be a compact or large region in the space. Other approaches are multilayer perceptron \cite{brown,Seow,Phung2}, Bayesian classifiers \cite{Sebe,Friedman,jones} and random forest \cite{random}. 

These aforementioned solutions that use single features, although successfully applied to human skin detection; they still suffer from: \textbf{Low Accuracy} False skin detection is a common problem when there is a wide variety of skin colours across different ethnicity, complex backgrounds and high illumination in image(s). \textbf{Luminance-invariant space} Some robustness may be achieved via the use of luminance invariant colour space \cite{Vadakkepat,Ukil}, however, such an approach can withstand only changes that skin colour distribution undergo within a narrow set of conditions and also degrades the performance \cite{colortrans}. \textbf{Require large training sample} In order to define threshold value(s) for detecting human skin, most of the state-of-the-art work requires a training stage. One must understand that there are tradeoffs between the size of the training set and classifier performance. For example, \cite{jones} required 2 billion pixels collected from 18,696 web images to achieve optimal performance. 

In this paper, we propose a novel approach - fusion framework that uses product rules on two features; the smoothed 2D histogram and Gaussian model to perform automatic skin detection. First of all, we employ an online dynamic approach as in \cite{pratheepan} to calculate the skin threshold value(s). Therefore, our proposed method does not require any training stage beforehand. Secondly, a 2D histogram with smoothed densities and a Gaussian model are used to model the skin and non-skin distributions, respectively. Finally, a fusion strategy framework using the product of two features is employed to perform automatic skin detection. To the best of our knowledge, this is the first attempt that employs a fusion strategy to detect skin in colour image(s).

The image pixels representation in a suitable colour space is the primary step in skin segmentation in colour images. A better 
survey of different colour spaces (e.g. RGB, YCbCr, HSV, CIE Lab, CIE Luv and normalised RGB) for skin colour representation and skin-pixel segmentation methods is given by Kakumanu et al. \cite{kaku}. In our approach, we do not employ the luminance-invariant space. Indeed, we choose the log opponent chromaticity (LO) space \cite{fin}. The reasons are twofold: first, colour opponency is perceptually relevant as it has been proven that the human visual system uses an opponent colour encoding \cite{Ewald,Hurvich}; and secondly, in this LO colour space, the use of logarithms renders illumination change to a simple translation of coordinates. Most of the aforementioned solutions claimed that illumination variation is one of the contributing factors that degrade the performance of skin detection systems. However, our empirical results and \cite{colortrans} showed that the absents of luminance component does not affect the system performance.
The remainder of the paper is structured as follows: Section \ref{Literature} gives a brief description of related work in human skin segmentation. Section \ref{methodology} derives our proposed fusion strategy. Section \ref{results} presents the experimental results using three different datasets and Section \ref{end} concludes the paper with discussions and future work. 
\section{Related Work}
\label{Literature}
Skin detection is the process of finding skin-colour pixels and regions in an image or video. In images and videos, skin colour is an indication of the existence of humans in media. In one of the early applications, detecting skin-colour regions was used to identify nude pictures on the Internet for content filtering. In another early application, skin detection was used to detect anchors in TV news videos for the sake of video automatic annotation, archival, and retrieval. Interested readers are encouraged to refer to \cite{kaku, Mitra} for a detailed background review.

\begin{figure}[t]
\centering
\includegraphics[scale=0.34]{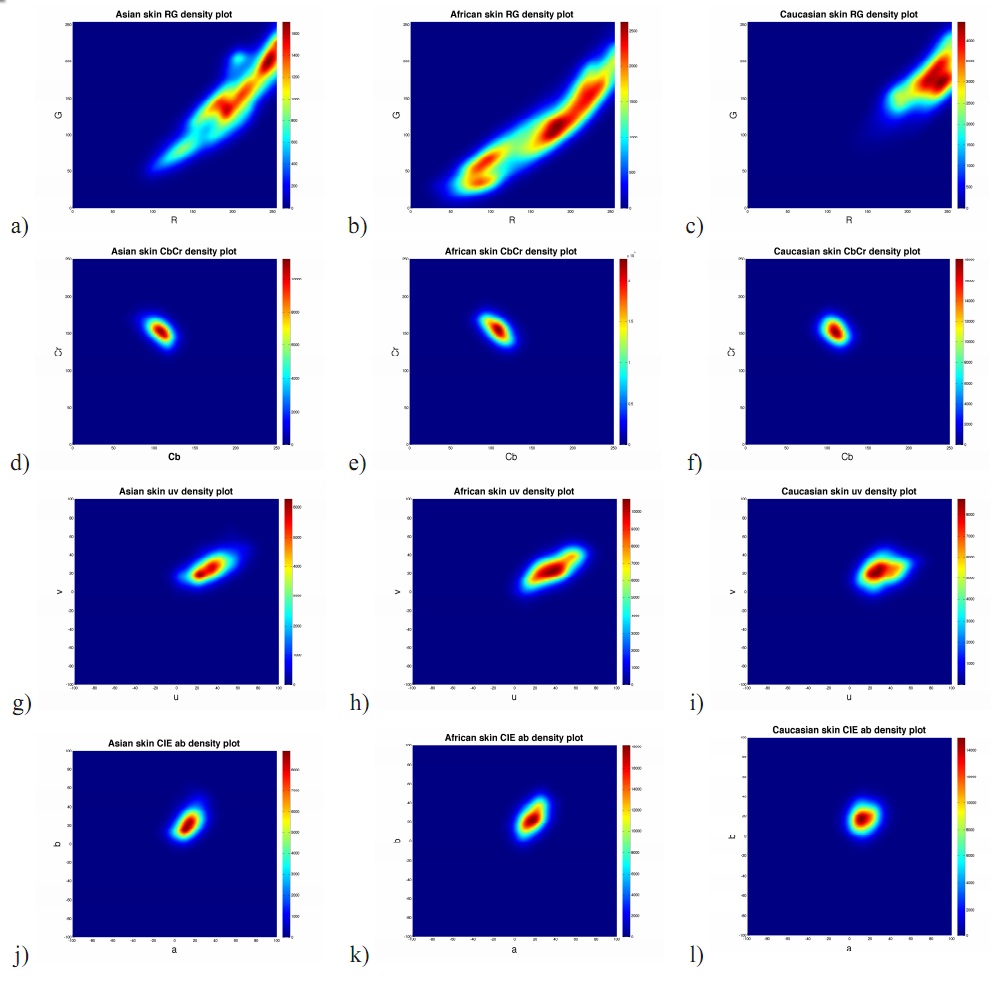}
\caption{Density plots of Asian, African and Caucasian skin in different colour spaces, adopted from Elgammal et al. \cite{Elgammal}. Each row represents different colour space and columns from left to right represent Asian, African and Caucasian, respectively}
\label{fig:skinp}
\end{figure}

A skin detector typically transforms a given pixel into an appropriate colour space and then uses a skin classifier to label the pixel whether it is skin or non-skin. A skin classifier defines a decision boundary of the skin colour class in the colour space based on a training database of skin-colour pixels. For example, Sobottka and Pitas \cite{Sobottka} used fixed range values on the $HS$ colour space where the pixel values belong to skin pixels in the range of $R_H$ = [0, 50] and $R_S$ = [0.23, 0.68]. Wang and Yuan \cite{Wang} used threshold values in $RG$ space and $HSV$ space where threshold values are set to be within the range $R_r$ = [0.36, 0.465], $R_g$ = [0.28, 0.363], $R_H$ = [0, 50], $R_S$ = [0.20, 0.68] and $R_V$ = [0.35, 1.0] to differentiate skin and non-skin pixels. In these approaches, high false skin detection is a common problem when there are a wide variety of skin colours across different ethnicity, complex backgrounds and high illumination. Fig. \ref{fig:skinp} shows that the skin colour of people belongings to Asian, African, Caucasian groups is different from one another and ranges from white, yellow to dark. Some robustness may be achieved via the use of luminance invariant colour spaces \cite{Vadakkepat,Ukil}, however, such an approach can only cope if the change in skin colour distribution is within a narrow set of conditions \cite{colortrans}.

\begin{figure*}[ht]
\centering
\includegraphics[scale=0.43]{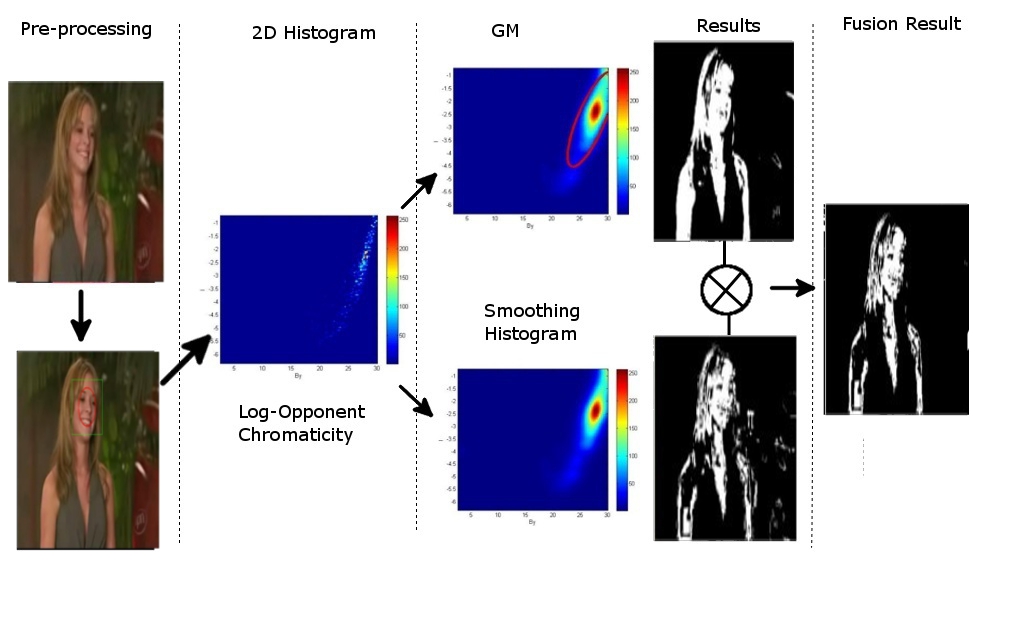}
\caption{The proposed framework: Eye detector, 2D Histogram, Gaussian model, Fusion Strategy}
\label{fig:flow}
\end{figure*}

Other approaches are multilayer perceptron \cite{brown,Seow,Phung2}, Bayesian classifiers \cite{Sebe,Friedman,jones} and random forest \cite{random}. In multilayer perceptron based skin classification, a neural network is trained to learn the complex class conditional distributions of the skin and non-skin pixels. Brown et al. \cite{brown} proposed a Kohonen network-based skin detector where two Kohonen networks; skin only and skin plus non-skin detectors were trained from a set of about 500 manually labelled images to obtain an optimal result. Sebe et al. \cite{Sebe} used a Bayesian network with training data of 60,000 samples for skin modelling and classification. Friedman et al. \cite{Friedman} proposed the use of tree-augmented Naive Bayes classifiers for skin detection. The Bayesian decision rule to minimum cost is a well-established technique in statistical pattern classification. Jones and Rehg \cite{jones} used the Bayesian decision rule with a 3D $RGB$ histogram model built from 2 billion pixels collected from 18,696 web images to perform skin detection. Readers are encourage to read \cite{kaku, Mitra} for a detailed state of the art review. Although these solutions had been very successful, they suffer from a tradeoff between precision and computational complexity.

In summary, our proposed method has two advantages in comparison to the state-of-the-art solutions. First of all, our proposed skin detection method employs an online dynamic threshold approach. With this, a training stage can be eliminated. Secondly, we select a fusion strategy for our skin detector. To the best of our knowledge, this is the first attempt that employs a fusion strategy to detect skin in colour image(s).
\section{Our Method}
\label{methodology}

Fig. \ref{fig:flow} shows the proposed framework for automatic skin detection. First, an approach similar to Fusel et al. \cite{eyedetect} is adopted to obtain the face(s) in a given image. Secondly, a dynamic method is employed to calculate the skin threshold value(s) on the detected face(s) region. Thirdly, two features - the 2D histogram with smoothed densities and Gaussian model are introduced to represent the skin and non-skin distributions, respectively. Finally, a fusion framework that uses the product rule on the two features is employed to obtain better skin detection results. In this paper, the RGB colour space is converted to the log opponent chromaticity space \cite{fin} to mimic visual human perception \cite{Ewald}.
\subsection{Pre-processing}
\label{preprocess}

\begin{figure}[htbp]
\centering
\includegraphics[scale=0.48]{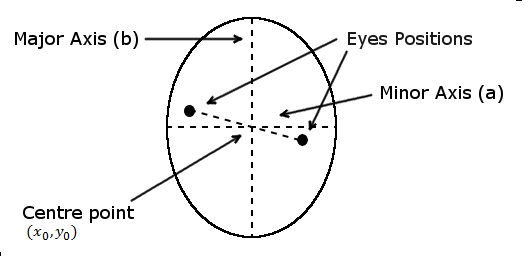}
\caption{Elliptical mask model generated using eye coordinates}
\label{fig:mask}
\end{figure}

In the pre-processing steps, for any given image(s), $S_t$ where $t$ is the number of images, $t \in \{1,2,\cdots,T\}$ we first locate human eyes as Fusel et al. \cite{eyedetect}. Then, an elliptical mask model as illustrated in Fig. \ref{fig:mask} is used to generate the elliptical face region in the image(s). Here, $(x_0, y_0)$ is the centre of the ellipse as well as the eyes symmetry point. Minor and major axes of the ellipse are represented by $1.6D$ and $1.8D$ respectively, where $D$ is the distance between two eyes. For a more detailed description, interested readers are encouraged to read \cite{eyedetect}.

The detected face regions include smooth (i.e. skin) and non-smooth (i.e. eyes, eye brown, mouth etc.) textures. As we are only interested in smooth regions, Sobel edge detection is employed to remove non-smooth regions. The choice of Sobel edge detection method is due to computational simplicity. Then, the detected edge pixels are further dilated using a dilation operation to get the optimal non-smooth regions. Finally, we obtain a new image(s), $S{'_t}$ that only consist(s) of face regions. 
\subsection{Colour Space}
\label{LOG}

It is well established that the distribution of colours in an image is often a useful cue. An image can be represented in a number of different colour space models (i.e. $RGB$, $HSV$ \cite{Vadakkepat}, $YC_rC_b$ \cite{Kumar}). These are some colour space models available in image processing. Therefore, it is important to choose the appropriate colour space for modelling human skin colour.

In this paper, we propose the use of the log opponent chromaticity colour space \cite{Forsyth}, the reason is twofold: first, colour opponency is perceptually relevant as it has been proved that the human visual system uses an opponent colour encoding \cite{Ewald,Hurvich}; and secondly, in this colour space, the use of logarithms renders illumination change to a simple translation of coordinates. 
\subsubsection{Log Opponent Chromaticity Space}

The theory of opponent colours was first studied by Hering \cite{Ewald} in 1892. He observed that certain colours are never perceived together in the human visual system. For instance, we never see yellowish-blue or reddish-green. Based on this theory, the log opponent chromaticity (LO) is a representation of colour information by applying logarithms to the opponency model so that it is simple to model illumination changes. As illumination changes, log component chromaticity distributions undergo a simple translation. These distributions are coded by using means and first $k$-moments found using principle component analysis \cite{fin}. 

\subsection{Skin Detection}
\subsubsection{Dynamic threshold with smoothed 2D histogram}

Human skin colour varies greatly between different ethnicity \cite{Elgammal}. Nonetheless, skin appearance in colour image(s) can also be affected by illumination, background image, camera characteristic etc. Therefore, a fixed or pre-learned threshold for detecting skin boundaries is not a feasible solution. 

\begin{figure}[htbp]
\centering
\subfigure[1D Histogram]{
\includegraphics[scale=0.25]{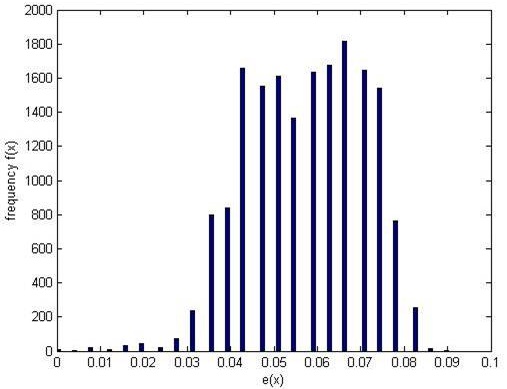}
\label{fig:1dhist}}
\subfigure[2D Histogram]{
\includegraphics[scale=0.30]{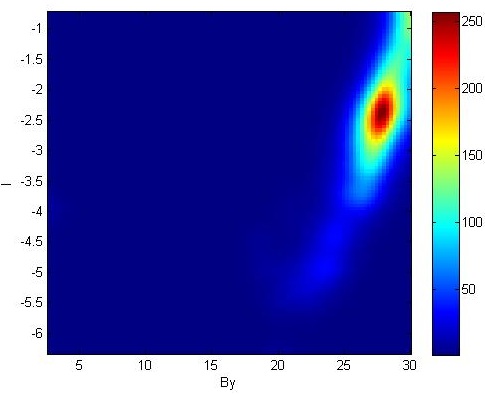}
\label{fig:2dhist}}
\caption{Histograms of 1D and 2D. Fig. \ref{fig:1dhist} has only one channel with frequency at $y-axis$. Fig. \ref{fig:2dhist} has 2 different channels of the same colour space on $x-axis$ and $y-axis$.}
\label{fig:hist}
\end{figure}

In our approach, we employ an online dynamic approach as to \cite{pratheepan} to calculate the skin threshold value(s) on the face images, $S{'_t}$. The assumption is that the face and body of a person always share the same colours. However, instead of using the 1D histogram as illustrated in Fig. \ref{fig:1dhist}; we introduce a 2D histogram (Fig. \ref{fig:2dhist}) with smoothing densities \cite{eilers}. 

In this paper, the feature vector for the smoothed 2D histogram, $Z$ is represented by the combination of $I$ and $B_y$. The smoothed 2D histogram-based skin segmentation, $D_{hist}$, at pixel $n$ is given as: 
\begin{equation}
\begin{array}{l l l}
D_{hist}(S_t, Z) &=& \left\{
\begin{array}{l l}
1 & \text{if $Z(I_n, By_n) > 20$}\\
0 & \text{if $Z(I_n, By_n) \leq 20$}
\end{array} \right.\\
&& ,\text{$\forall I_n, By_n \in S_t$}
\end{array}
\label{eq:dhist}
\end{equation}
\subsubsection{Gaussian Model}

The Gaussian model is a sophisticated model that is capable of describing complex-shaped distributions and is popular for modelling skin colour distributions. The threshold skin colour distribution in the 2D histogram is modelled through elliptical Gaussian joint probability distribution functions defined as: 

\begin{equation}
p(H|\lambda) = \sum^{k}_{i=1}\omega_i g(c|\mu_i, \Sigma_i)
\label{eq:gmm}
\end{equation}
\noindent where $H$ is the colour vector of $(I, B_y)$, $\lambda = \{\omega_i, \mu_i, \Sigma_i\}$ , $\mu_i$ is the mean vector and $\Sigma_i$ is the diagonal covariance matrix, respectively. $\omega_i$ refers to the mixing weights, which satisfy the constraint $\sum^{k}_{i=1}\pi_i = 1$. 
\begin{figure}[htp]
\centering
\includegraphics[scale=0.5]{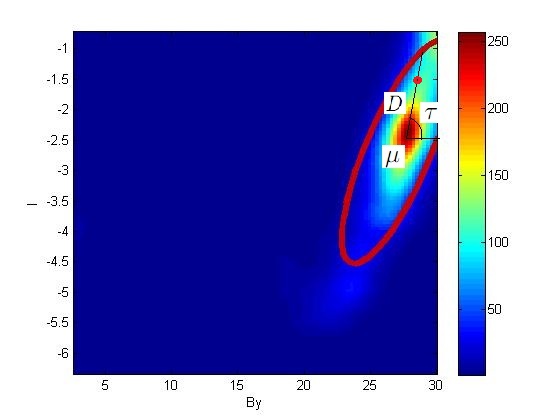}
\caption{Gaussian Model}
\label{fig:gmmCalc}
\end{figure}

The result of Gaussian model-based skin detection, $D_{gmm}$, can be obtained by using Fig. \ref{fig:gmmCalc}. $\mu$ is the center of the Gaussian model, while $\tau$ is the angle between x-axis and line $D$. Let $(I_i, By_i)$ be the coordinate of pixel $n$ and is position on the red dot along line $D$. Distance of $(I_n, By_n)$, $d$ and angle, $\tau$ are calculated as Eq. \ref{eq:distanced} and \ref{eq:tau}.

\begin{equation}
d = \sqrt{d_x^2 + d_y^2}
\label{eq:distanced}
\end{equation}
\begin{equation}
\tau = tan^{-1}(\frac{d_y}{d_x})
\label{eq:tau}
\end{equation}
\noindent $d_x$ and $d_y$ are the distances between $(I_n, By_n)$ and center, $\mu$ at x-axis and y-axis, respectively. $\mu_x$ and $\mu_y$ are the coordinate of $\mu$ at x-axis and y-axis, respectively. Distance between the boundary and center of the Gaussian model at x-axis and y-axis, $D_x$ and $D_y$ at given angle, $\tau$ are:

\begin{equation}
D_x = \Sigma_x cos(\tau) 
\label{eq:xD}
\end{equation}
\begin{equation}
D_y = \Sigma_y sin(\tau)
\label{eq:yD}
\end{equation}

\noindent where $\Sigma_x$ and $\Sigma_y$ are the variance of x-axis and y-axis for Gaussian model. Distance, $D$ is represented as:

\begin{equation}
D = \sqrt{D_x^2 + D_y^2}
\label{eq:distanceD}
\end{equation}

Therefore, $D_{gmm}$ is given:
\begin{equation}
D_{gmm}(S_t, \mu, \Sigma) = \left\{
\begin{array}{l l}
1 & \text{if $D > d$} \\
0
\end{array} \right.
\label{eq:dgm}
\end{equation}
\subsection{Fusion Strategy}
\label{fusion}

In order to increase the effectiveness and robustness of the skin detection algorithm, a fusion strategy is proposed by integrating the two incoming single features into a combined single representation. Both modals will vote for classification of skin and non-skin pixels. This can be done by using product rule to both modals. Let $D_{hist}(S_t, Z)$, and $D_{gmm}(S_t, \mu, \Sigma)$ denote the matching results produced by the smoothed 2D histogram, $Z$ and Gaussian model, $g(c|\mu_i, \Sigma_i)$ respectively. The combined matching results $D(S_t)$ using the fusion rules can be obtained as the following:

\begin{equation}
D(S_t) = \Gamma\{D_{hist}(S_t, Z), D_{gmm}(S_t, \mu, \Sigma)\}
\label{eq:fusion}
\end{equation}

\noindent where $\Gamma$ is the selected fusion rule, which represents the product $\otimes$. In order to make the fusion issue tractable, the individual features are assumed to be independent of each other. 
\section{Experiments}
\label{results}

In this section, the performance of the proposed approach under different conditions, such as fusion strategy, colour spaces and a comparison with the state-of-the art methods in terms of qualitative and quantitative performance. We only perform quantitative analysis on the Stottinger dataset \cite{Julian} as ground truth videos are only available for this dataset.
\subsection{Datasets}

Experiments are conducted using three public databases: Pratheepan's dataset \cite{pratheepan}; ETHZ PASCAL dataset \cite{PASCAL} and the Stottinger dataset \cite{Julian}. The Pratheepan's dataset \cite{pratheepan} consists of a set of images downloaded randomly from Google. These random images are captured with a range of different cameras using different colour enhancements and under different illuminations. The ETHZ PASCAL dataset \cite{PASCAL} consists of 549 images from PASCAL VOC 2009. The dataset consists mainly of amateur photographs with difficult illuminations and low image quality; and finally the Stottinger dataset \cite{Julian} consists of popular public video clips from web platforms. These are chosen from the community (top-rated) and cover a large variety of different skin colours, illuminations, image quality and difficulty levels. 

\begin{figure}[htbp]
\centering
\includegraphics[scale=0.44]{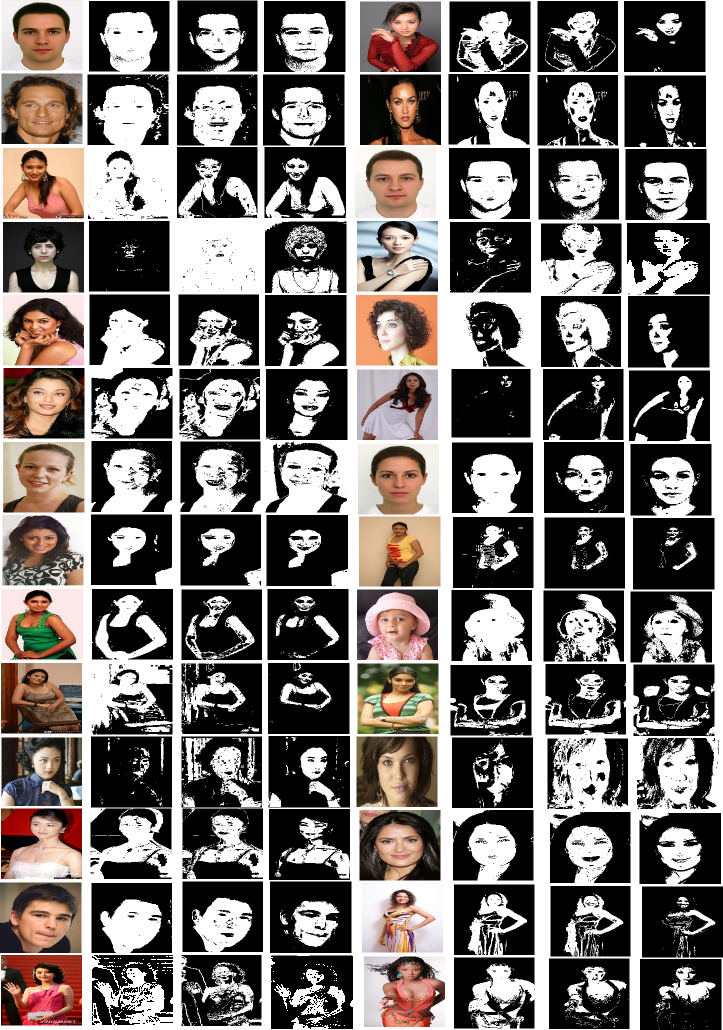}
\caption{Comparison using Pratheepan dataset \cite{pratheepan}. Column from left to right represent original images, \cite{cheddad} method, \cite{pratheepan} method, and our proposed methods respectively}
\label{fig:compare15}
\end{figure}
\subsection{Results and Analysis}

\begin{figure}[ht]
\centering
\includegraphics[scale=0.15]{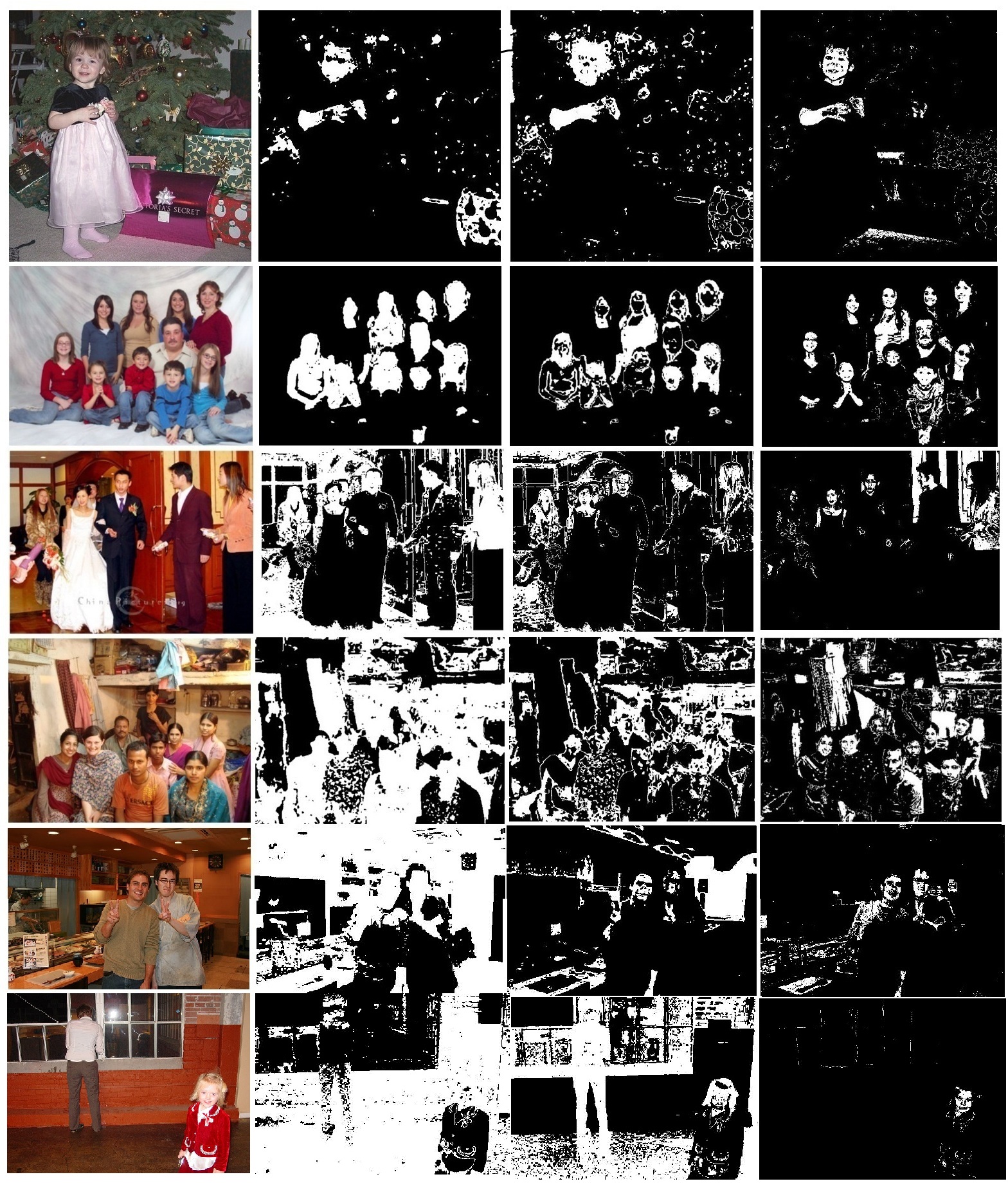}
\caption{Comparison using ETHZ dataset \cite{PASCAL}. Columns from left to right represent original images, \cite{cheddad} method, \cite{pratheepan} method, and our proposed method, respectively}
\label{fig:compare25}
\end{figure}

\begin{figure}[htbp]
\centering
\includegraphics[scale=0.23]{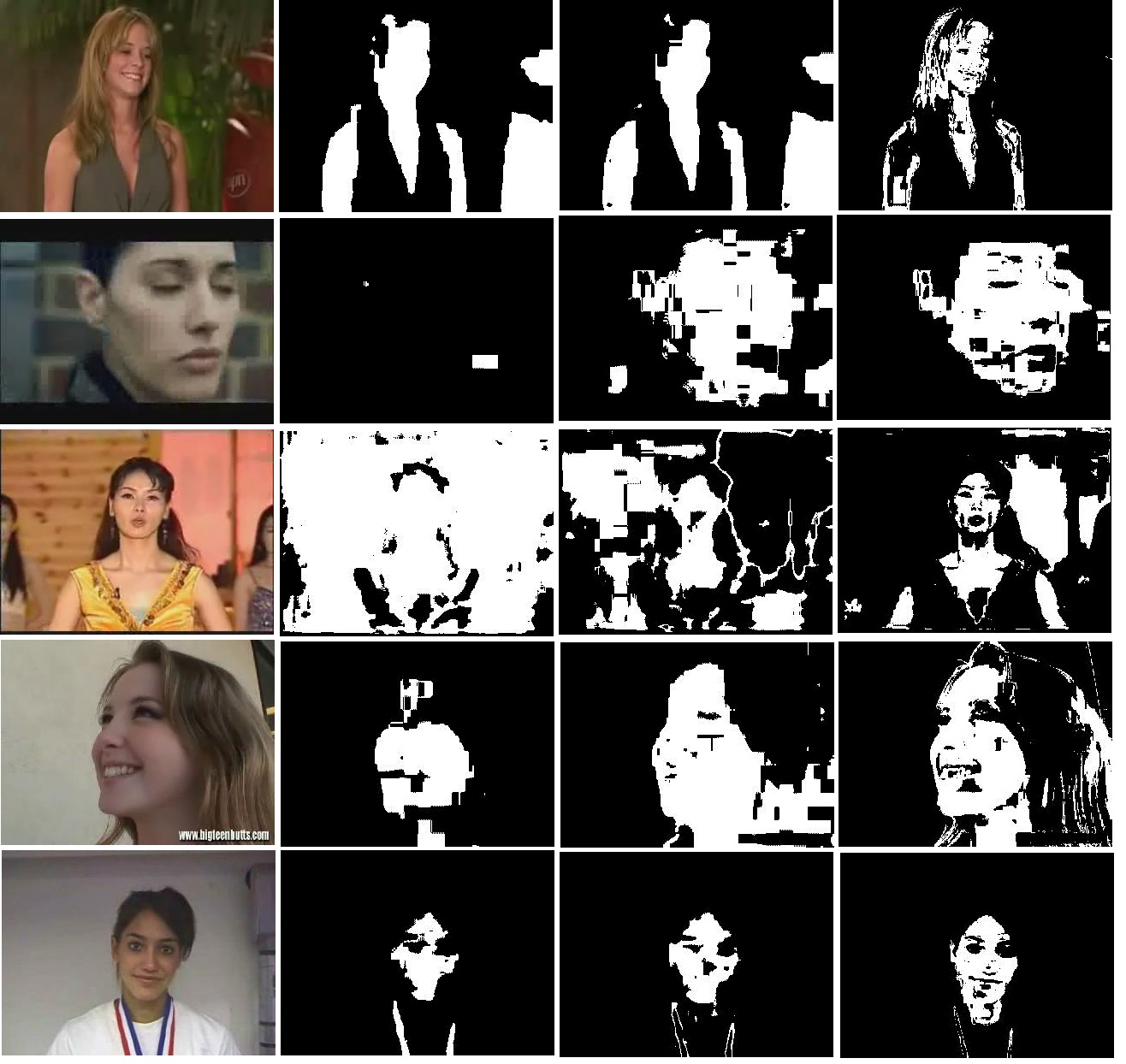}
\caption{Stottinger dataset \cite{Julian}. Columns from left to right represent original images, \cite{cheddad} method, \cite{pratheepan} method, and our proposed method respectively}
\label{fig:compareISVC}
\end{figure}

The detection results for each dataset are illustrated in Fig. \ref{fig:compare15} - \ref{fig:compareISVC}, respectively. When there is no face detected on the image, it will return a blank image (black). Therefore, for testing purposes, we assumed that true face(s) are detected in the image. Conclusions are drawn as follows; First, qualitatively, the proposed method has better detection accuracy in comparison to \cite{cheddad} and \cite{pratheepan}. For instance, in the image sample (row 4) of Fig. \ref{fig:compare15} and Fig. \ref{fig:compare25}, respectively, both the \cite{cheddad} and \cite{pratheepan} methods fail to detect the skin region accurately. One can notice from both images, that the particular image sample in Fig. \ref{fig:compare25} is a fairly complex environment with high illuminations and comprises of humans from different ethnicity. However, our proposed approach is still able to capture almost all skin regions and has the least noise. Secondly, the proposed approach has better robustness in terms of illumination, background image, camera characteristic and different ethnicity. The images in the Pratheepan dataset \cite{pratheepan} are captured with a range of different cameras using different colour enhancements. However, it can be qualitatively noticed that our proposed approach has the least effects on these factors. Nonetheless, in other datasets that are highly complex and illumination, our approach also achieves a better discrimination power than \cite{cheddad} and \cite{pratheepan}. Finally, the proposed approach does not require any training stage, and hence is more effective in terms of computational cost as opposed to approaches listed in \cite{random}. 

\subsection{Comparison between different colour spaces}

\begin{figure}[htp]
\centering
\includegraphics[scale=0.24]{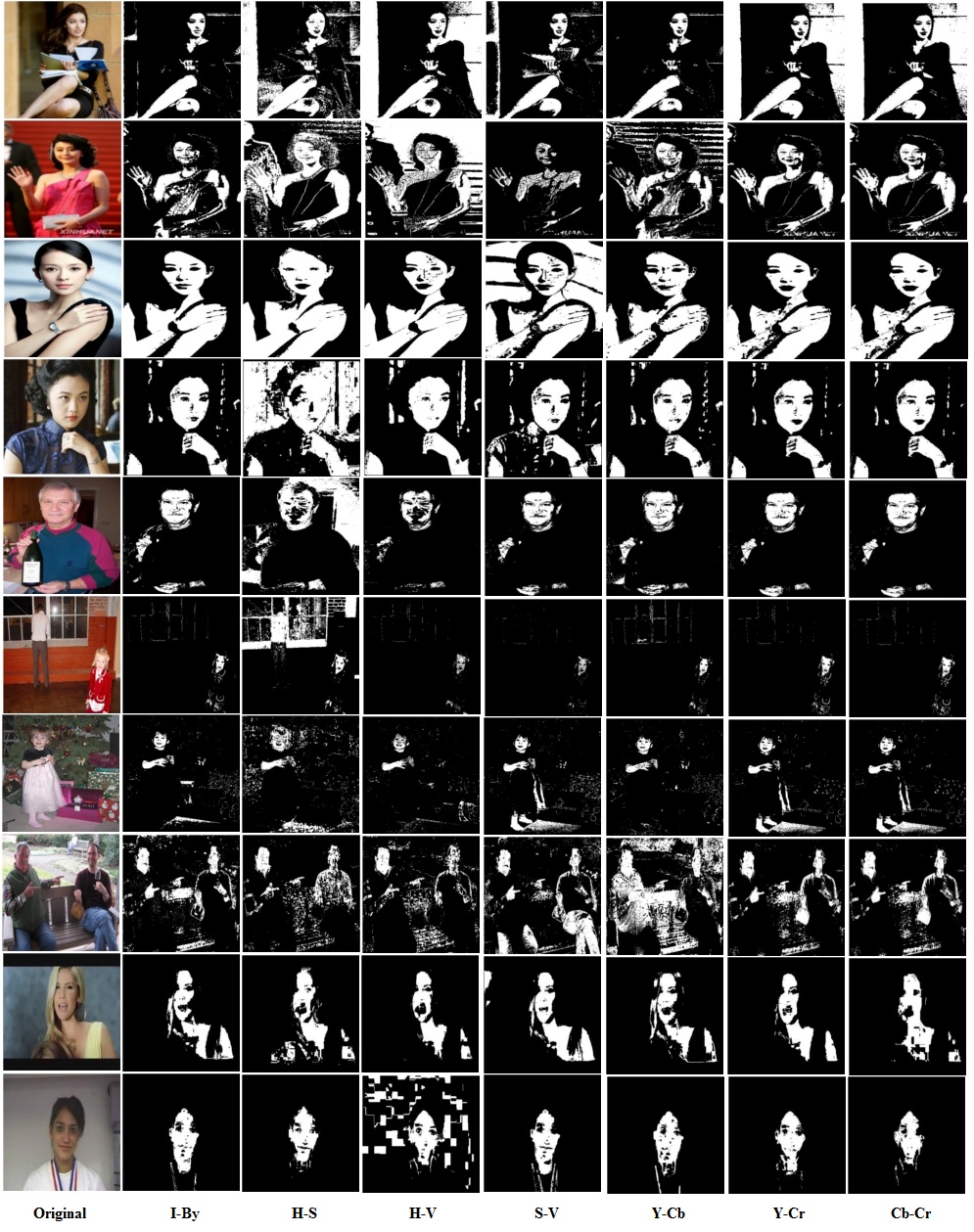}
\caption{Comparison between results from different colour space}
\label{fig:colorcomp}
\end{figure}

In this section, we analyse 7 different combinations of feature vectors: $IB_y$, $HS$, $HV$, $SV$, $YC_b$, $YC_r$ and $C_bC_r$. The results for each feature vector are presented in Fig. \ref{fig:colorcomp} using images from Pratheepan and ETHZ datasets. It can be noticed that $IB_y$ shows better true positive rate than the rest. Further, a better quantitative analyse result is shown in Table \ref{bStatistic} using Stottinger dataset as this dataset only contains ground-truth. It can be seen from Table \ref{bStatistic} that the results from $IB_y$ and HS are comparable. However we selected $IB_y$ as our colour space as it shows higher true positive rate and lower false negative rate than HS. Also it has been proven that the human visual system uses an opponent colour coding. 

\begin{table}[!t]
\renewcommand{\arraystretch}{1.3}
\caption{Comparison between Different Colour Spaces in Stottinger datasets \cite{Julian}}
\label{bStatistic}
\centering
\begin{tabular}{|c|c|c|c|c|}
\hline
& & & True & False \\
Colour Space & Accuracy & F-score & Positive & Negative \\
& & & Rate & Rate \\
\hline
\textbf{$IB_y$} & \textbf{0.9039} & \textbf{0.6490} & \textbf{0.6580} & \textbf{0.3420} \\
HS & 0.9057 & 0.6512 & 0.6521 & 0.3479\\
HV & 0.7977 & 0.4549 & 0.6251 & 0.3749\\
SV & 0.8898 & 0.6285 & 0.6905 & 0.3995\\
$YC_b$ & 0.8936 & 0.6143 & 0.6277 & 0.3723\\
$YC_r$ & 0.8985 & 0.6392 & 0.6656 & 0.3344\\
$C_bC_r$ & 0.9151 & 0.6241 & 0.5223 & 0.4777\\
\hline
\end{tabular}
\end{table}
\subsection{Fusion strategy results}

\begin{figure}[htbp]
\centering
\includegraphics[scale=0.4]{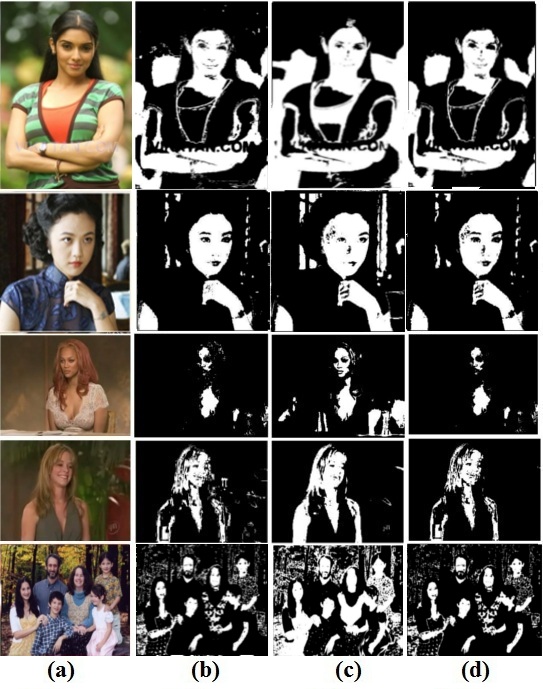}
\caption{(a) Original Image, (b) 2D Histogram's Result, (c) Gaussian model's result, and (d) Fusion approach's result}
\label{fig:fuse}
\end{figure}

In this section, we show the comparison results of using single feature - smoothed 2D histograms (s2D) or Gaussian mixture models (GMM) only, and multiple features (the fusion of s2D and GMM). The results are as illustrated in Fig. \ref{fig:fuse} and Table \ref{cStatistic}. Fusion Approach has the highest Accuracy and F-score. Moreover, it can also be visualised that the fusion strategy has lower false positive rate compare to the single feature approach. For instance, the smoothed 2D histogram is able to detect most of the skin regions; but it is highly occluded with noise. The Gaussian model in the meantime, has very high false positive rate (i.e. clothing which is classified as skin region). 

\begin{table}[!t]
\renewcommand{\arraystretch}{1.3}
\caption{Comparison between Our Proposed Fusion and Non-fusion Approach in $IB_y$ Colour Space using Stottinger dataset \cite{Julian}}
\label{cStatistic}
\centering
\begin{tabular}{|c|c|c|c|c|}
\hline
& & & True & False \\
Classifier & Accuracy & F-score & Positive & Positive \\
& & & Rate & Rate \\\hline
\textbf{Fusion} & \textbf{0.9039} & \textbf{0.6490} & \textbf{0.6580} & \textbf{0.0577}\\
s2D Histogram & 0.8930 & 0.6270 & 0.6662 & 0.0716\\
GMM & 0.8595 & 0.6150 & 0.8314 & 0.1361\\
\hline
\end{tabular}
\end{table}
\subsection{Quantitative Analysis}

\begin{figure}[htp]
\centering
\includegraphics[width=0.9\columnwidth]{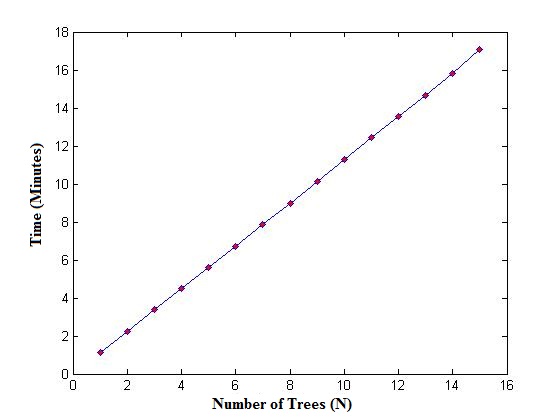}
\caption{Plot of time in minutes taken for 15 trees for offline training}
\label{fig:plot}
\end{figure}

Table \ref{statistic} shows comparisons between our method and other state-of-the-art methods on the Stottinger dataset \cite{Julian}. Apart from Random Forest \cite{random}, other methods do not include training. a total of 2985 frames were extracted from 7 videos for our experiment. For Random Forest \cite{random}, 1990 image frames are randomly chosen for training and remaining images are used for testing. From those 1990 images, around 3 million pixels were randomly chosen and 15 tress were trained. Each tree extracts 70\% of the pixels respectively for training. Figure \ref{fig:plot} shows the time in minutes required for 15 trees to be trained in random forest solutions. It took around 17 minutes for 15 trees to be trained. 

\begin{figure}[htbp]
\centering
\includegraphics[width=0.85\columnwidth]{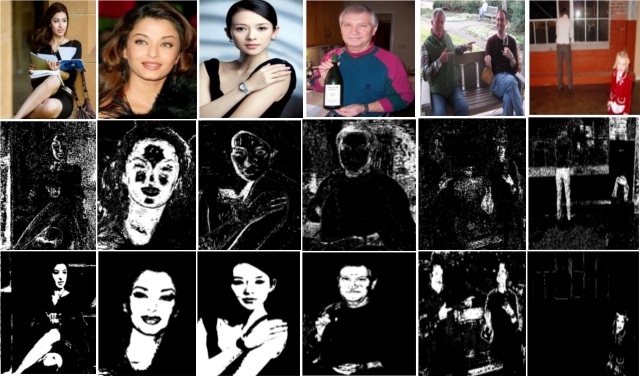}
\caption{Comparisons between random forest and our method using Pratheepan dataset\cite{pratheepan} and ETHZ dataset \cite{PASCAL}. Row from top to bottom represent original image, result from random forest and result from our method, respectively}
\label{fig:rferror}
\end{figure}

Figure \ref{fig:rferror} shows the comparisons between random forest and our method using the Pratheepan dataset\cite{pratheepan}. Here Random Forest \cite{random} is trained with Stottinger dataset and tested with Pratheepan dataset\cite{pratheepan}. It shows that random forest does not work well. In order to increase the accuracy of random forest on this dataset, huge training samples and/or more trees will be needed. This will cause higher computational power as number trees increases and time consuming during training. Our method is able to maintain quality on skin segmentation.

\begin{table}[t]
\renewcommand{\arraystretch}{1.3}
\caption{Comparison between Classifier Performance in Stottinger dataset \cite{Julian} using $IB_{y}$ colour space}
\label{statistic}
\centering
\resizebox{!}{1.1cm}{
\begin{tabular}{|c|c|c|c|c|}
\hline
Classifier & Accuracy & F-score & Precision & Recall\\
\hline
Random Forest \cite{random} & 0.9305 & 0.7307 & 0.7661 & 0.6984\\
\textbf{Our Proposed Method} & \textbf{0.9039} & \textbf{0.6490} & \textbf{0.6403} & \textbf{0.6580}\\
Dynamic Threshold \cite{pratheepan} & 0.8935 & 0.5922 & 0.6133 & 0.5725\\
Static Threshold \cite{cheddad} & 0.8334 & 0.4745 & 0.4133 & 0.5570\\
\hline
\end{tabular}}
\end{table}
A quantitative analyse is shown in Table \ref{statistic} using Stottinger dataset. The accuracy is a score that uses the sum of true positive and true negative as measurement. While, $F\textrm{-}score$ is the product of $Precision$ and $Recall$. Table \ref{statistic} shows that our proposed method has an acceptable score as compared to Random Forest \cite{random} even no training stage is required. Nonetheless, it is able to cope with large variable of illumination and complex backgrounds variations. 
\begin{figure}[tp]
\centering
\includegraphics[scale=0.2]{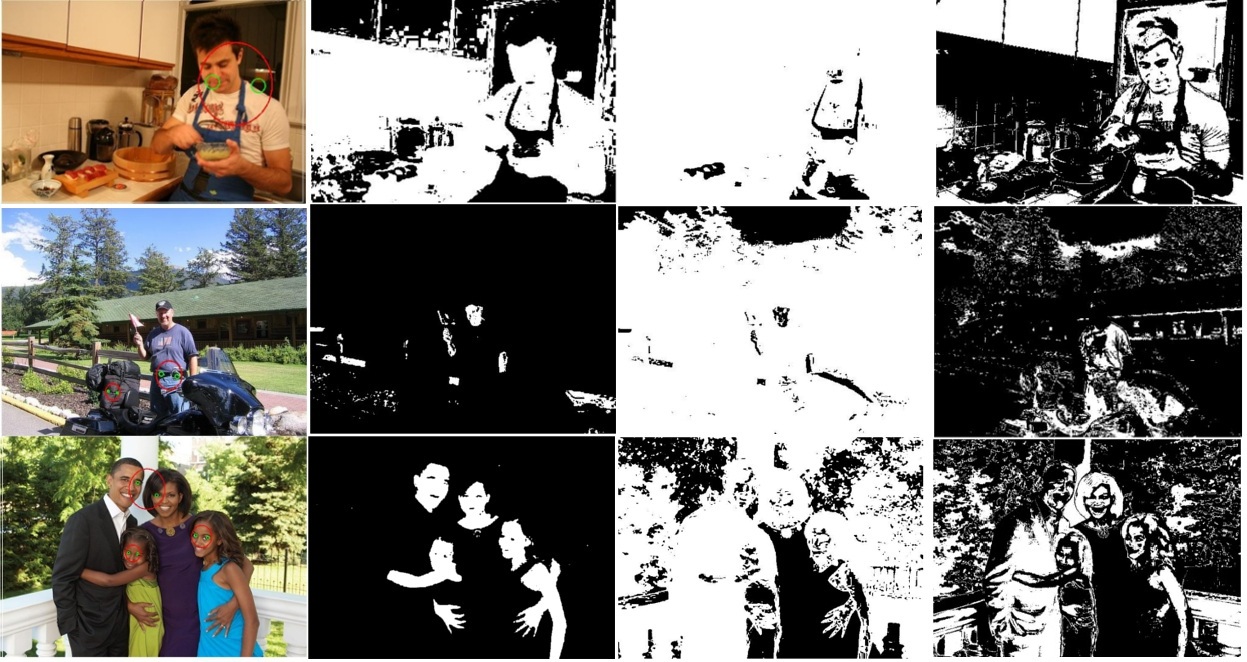}
\caption{Column from left to right represent original images, skin detection using \cite{cheddad}, \cite{pratheepan} and our proposed method respectively}
\label{fig:bad}
\end{figure}

\subsection{Discussions}

In our experiments, we noticed that the final result of our work depends greatly on the outcome of the pre-processing phase (the eye detector \cite{eyedetect}). If the algorithm detects a false face region, poor result will be returned. Fig. \ref{fig:bad} shows the result of skin segmentation for false face region detected. When a false face region is obtained, false dynamic thresholds will be generated. Therefore, false classifications will be processed, where non-skin regions are classified as skin regions. In our future work, we will investigate the face detector algorithm to overcome this problem.
\section{Concluding Remarks}
\label{end}

In this paper, a fusion framework based on smoothed 2D histogram and Gaussian model has been proposed to automatic detect human skin in image(s). As exhibited in experiments, the proposed method outperforms state-of-the-art methods in terms of accuracy in different conditions: background model, illumination and ethnicity. With this, it shows the potential to be applied to a range of applications such as gesture analysis. One drawback of the proposed approach is that its success relies on eye detector algorithms. However, this is the general problem faced by all other researchers who work in this domain. Our future work is focused on building a better pre-processing method; to use field programmable gate arrays to implement a hardware scheme.
%
\bibliographystyle{IEEEtran}
\bibliography{tran}

\end{document}